# Human-AI Collaboration: From Paradoxes to Patterns*

Michael Weiss

Carleton University, Ottawa, Canada

michael_weiss@carleton.ca

## Abstract

Evidence shows that humans and AI systems perform better together, by collaborating, than alone. This paper examines two key design dimensions of human-AI collaboration (autonomy and initiative) and explores the collaboration patterns that they generate. Documenting these patterns starts with identifying the underlying problems and solutions, followed by examining the internal tensions within the problems. The paper uses a paradox perspective to analyze those tensions. It describes a process for surfacing the tensions and mapping the underlying paradoxes. It also illustrates how the pattern descriptions can be derived from mapping these paradoxes. Finally, the paper documents four human-AI collaboration patterns: Instruction, Delegation, Assistance, and Co-creation.

## 1 Introduction

Evidence shows that humans and AI systems perform better together, by collaborating, than alone. The canonical example is a doctor combining their expertise with the predictions of an AI model. This paper examines two key design dimensions for human-AI collaboration (autonomy and initiative) and explores the collaboration patterns that they generate. The focus is on the centaur model, in which the human is in control, rather than the reverse centaur model (Doctorow, 2026).

Documenting these patterns starts with identifying the underlying problems and solutions, followed by examining the internal tensions within the problems. The paper uses a paradox perspective to analyze those tensions (Weiss, 2026). Formally, a paradox is a tension between contradictory yet interdependent forces whose resolution recreates the very conditions that caused it (Smith & Lewis, 2011).

The paper describes a process for surfacing the tensions and mapping the underlying paradoxes. Furthermore, it illustrates how the pattern descriptions can be derived from mapping the paradoxes.

---

The audience for these patterns includes anyone who wants to collaborate with AI more effectively, as well as pattern authors who want to learn how to begin writing their patterns from a paradox.

The paper is structured as follows. The next section provides background on reframing tensions as paradoxes and human-AI collaboration. This is followed by an exploration of the design dimensions for human-AI collaboration and the resulting design space. Next, the paper introduces a process for mapping tensions to paradoxes and deriving pattern descriptions from these mappings. The descriptions of the patterns follows. The paper closes with a summary of contributions and future work.

# 2 Background and related work

## 2.1 Reframing tensions as paradoxes

A core element of each pattern is a description of the forces that make the problem challenging and the tensions that exist between them. In the literature, forces are often viewed as competing rather than complementary. Such an either/or framing asks us to choose one side of a tension at the expense of the other side. While this may be the best way to address certain tensions, often a more holistic and sustainable way is to embrace both sides of the tension (both/and framing).

A both/and framing leads to a view of tensions as paradoxes. According to Smith & Lewis (2011), a tension is paradoxical when its two sides are contradictory yet interdependent, and the tension persists whenever we try to resolve it. In fact, a paradoxical tension cannot be resolved once and for all, only managed or navigated. Take the example of building a software system:

Common advice is to make a software system modular to make it easier to maintain. However, the initial version of a system is usually highly integrated, because speed of development matters more at this stage than maintainability. As the system grows, however, maintainability takes a higher priority. This creates a tension: integrated vs modular. To manage this tension, system designers first choose one, then the other side of the tension, repeating this cycle over the system's lifetime.

Weiss (2026) gives examples of paradoxical tensions in existing patterns. This paper introduces a circular response template for organizing the tensions and strategies to manage them. It consists of circles for the opposing forces of a tension (A and B), the outcomes associated with these forces, and arrows indicating the pull from the opposing forces, as well as the type of strategy. The strategies are either to alternate between the forces (differentiation) or address them at the same time (integration).

## 2.2 Human-AI collaboration

As AI becomes more capable and pervasive, it is changing from being a mere tool to turning into a partner (Lindgren, 2025). This changes how we perceive our relationship towards AI: from using AI to collaborating with AI. In the tool-based model, AI follows instructions in a deterministic, linear and unambiguous manner. In the collaborative model, AI acts as an active collaborator, rather than a passive instrument (Salma et al., 2025). This model matches the dynamic, non-linear, and ambiguous nature of complex tasks that require creative approaches. This is not a gradual change, but a shift in how agency and responsibility are distributed between humans and AI.

There is not much existing research on human-AI collaboration patterns. Pautasso & Xu (2026) is one of the few papers to document patterns in this space. They describe ten patterns grouped into three categories: basic, sequential, and parallel. In the basic category, a fundamental distinction is made in terms of who drives the interaction: human (Centaur) or AI (Reverse Centaur). The remaining patterns assume that human and AI collaborate to process a given input. The patterns are grouped in two basic flow structures: whether human and AI process input in parallel or in sequence. The key design dimensions the authors consider include responsibility of input and output, flow, conflict resolution, autonomy, and initiative. The last two dimensions form the core of the design space for this paper.

Papers that discuss paradox in the context of AI include Raisch & Krakowski (2021) and Salma et al. (2025). For Raisch & Krakowski (2021), the central paradox is between automation and augmentation. Automation occurs when AI takes over a task, augmentation when humans collaborate closely with AI on the task. Automation and augementation are contradictory, yet interdependent. To understand this, consider that automation is only possible if a task is well-understood and robust. Initially, humans need to work closely with AI (augmentation) to refine the models so they can be automated. However, when conditions change, automated solutions fail and augmentation becomes necessary again.

Salma et al. (2025) identified five underlying paradoxes in their study of human-AI collaboration for creative tasks: ambiguity vs. precision, control vs. serendipity, speed vs. reflection, individual vs. collective, and originality vs. remix. These are core tensions that inform the design of human-AI co-creative systems. They are fundamental to the nature of human-AI collaboration, and cannot be eliminated by making the "right" design decisions. Take, for instance, ambiguity vs. precision. This tension reflects the fact that human intent is vague while AI requires precise input. This translates into a design goal to enable users to represent their vision in prompts without limiting their ability to explore.

# 3 Design dimensions for human-AI collaboration

Two key design dimensions are the degree to which the AI system (or AI for short) operates independently (autonomy) and how often it takes the initiative. The first dimension coincides with the scope of the task assigned to the AI, in other words, whether it is kept on a short or long leash. The second dimension captures whether the AI waits for human input or acts on its own.

Together, these design dimensions give rise to four scenarios. Each requires a different combination of levels of autonomy and initiative. Figure 1 shows the design space for human-AI collaboration.

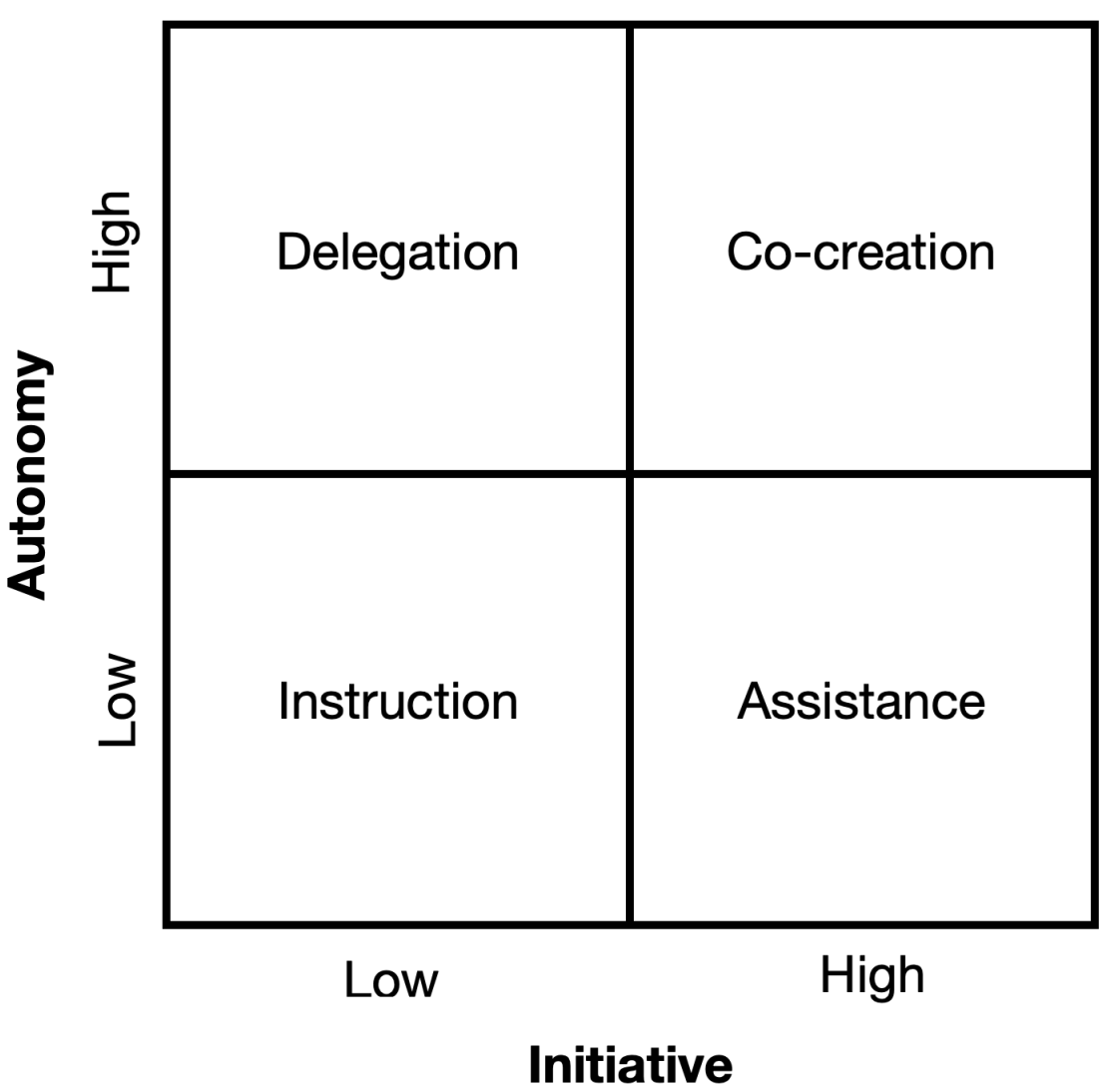

**Figure 1.** Design space for human-AI collaboration based on autonomy and initiative

How autonomy and initiative should be allocated depends on the context of the task to be accomplished. We can distinguish between four scenarios: the knowledge needed to perform the task is still tacit; the task is well-understood and self-contained; the task combines high-level cognitive work with low-level routine work; and the task requires multiple, diverse perspectives.

Table 1 cross-references the collaboration scenarios with these contexts.

**Table 1.** Collaboration scenarios and their contexts

| Scenario | Context |
|---|---|
| Instruction | The knowledge needed to perform the task is still tacit |
| Delegation | The task is well-understood and self-contained |
| Assistance | The task combines high-level cognitive work with low-level routine work |
| Co-Creation | The task requires multiple, diverse perspectives |

# 4 From tensions to paradoxes and patterns

## 4.1 Process of mapping a paradox

The circular response template (Weiss, 2026) only provides a static view of a paradox: the forces and the pull between the forces. It does not show how the paradox can be managed over time. To address this limitation, four elements of a paradox were identified (need, action and reaction, consequence, and circular response) in addition to the forces and the pull between them:

- Need: scenario-specific design constraints

- Action and reaction: response to address the need and reaction to the response
- Consequences: consequences of the action (the consequences of the reaction are symmetric)
- Circular response: cycle of action, consequence, reaction, consequence
- Tension: forces involved in the central tension and pull between the forces

The process of mapping a paradox starts with a need. This need is derived from the context. For example, in the instruction scenario, the context expresses the need to make tacit knowledge explicit, or to codify it. The next step is to identify how this need can be addressed, that is, through which pair of action and reaction. In the instruction scenario, the human writes prompts that instruct the AI and then observes the responses. Writing prompts is analogous to probing a complex system to see how it behaves, and observing the system causes the human to reflect on the observed behavior. Thus, a pair of action and reaction that describes how the human interacts with the AI is probe and reflect.

In probing the human articulates their tacit knowledge into prompts. To ensure that the prompts capture the user's intent, the user must monitor the AI's response for deviations from expected behavior. This is what Weick & Sutcliffe (2015) refer to as sensitivity of operations. At the same time, the structure of the prompt should initially be loosely defined so that it can be shaped as the user learns more about the AI's actual behavior. Dalsgaard (2025) refers to structuring as exploratory externalization; prompts evolve with feedback. These are the consequences of the probe action. Similarly, reflecting on the AI's response results in a revised or follow-up prompt that better captures the user's intent. However, this comes at the cost of sensitivity to what the new structure did not anticipate.

The circular response captures the cycle of actions. It documents how the paradox persists, as the consequences of each action reinforce the next action. In the example scenario, the cycle starts with the probe and its consequences (gaining sensitivity and losing structure) and continues with reflection and its consequences (losing sensitivity and gaining structure). The final element of the template is the tension. It consists of two central forces that are on opposite poles of the tension. Here, the poles refer to the tacit and explicit nature of knowledge. The labels on the arrows between the tensions show the actions that pull from one force to the other. For example, probing pulls from tacit to explicit.

While this suggests a linear order of steps to map a paradox, the actual process is iterative.

## 4.2 Paradoxes in human-AI collaboration scenarios

Figure 2 shows the mapping of the paradox for the instruction scenario. This scenario is driven by the need to codify tacit knowledge. In order to codify their knowledge, a human user probes an AI by prompting it. The consequence of prompting the AI are high sensitivity (understanding of the AI's behavior) and loose structure (the logic of the prompt is loosely defined). The corresponding reaction is to reflect on the AI's response (noticing deviations from the user's intent and reflecting them in an improved prompting logic). The central tension underlying the paradox is between tacit and explicit knowledge. Probing and reflection combine to form a circular response to this tension.

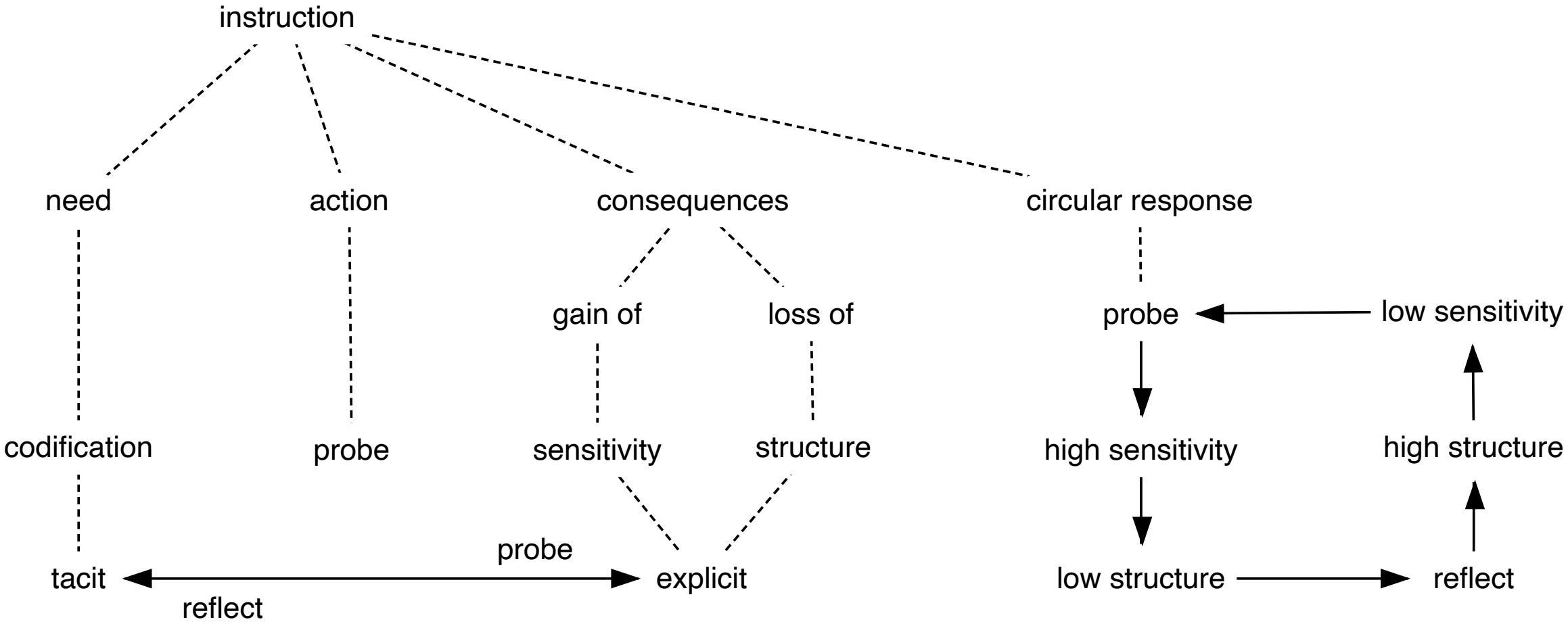

**Figure 2.** Mapping of the paradox for the instruction scenario.

The delegation scenario in Figure 3 serves to increase the human user's capacity to get work done and reallocate their focus to higher-priority tasks by offloading well-understood and clearly scoped tasks to AI. The central paradoxical tension is between centralized and decentralized control. Offloading a routine task to AI decentralizes control and frees up the human to pursue cognitively more demanding tasks. The result of offloading is also a loss of oversight, though. The risk is that the human may just accept the output produced by AI, in other words, they disengage from how the output is produced. To balance this risk, it is critical to review (at least a sample of) the outputs by reclaiming control.

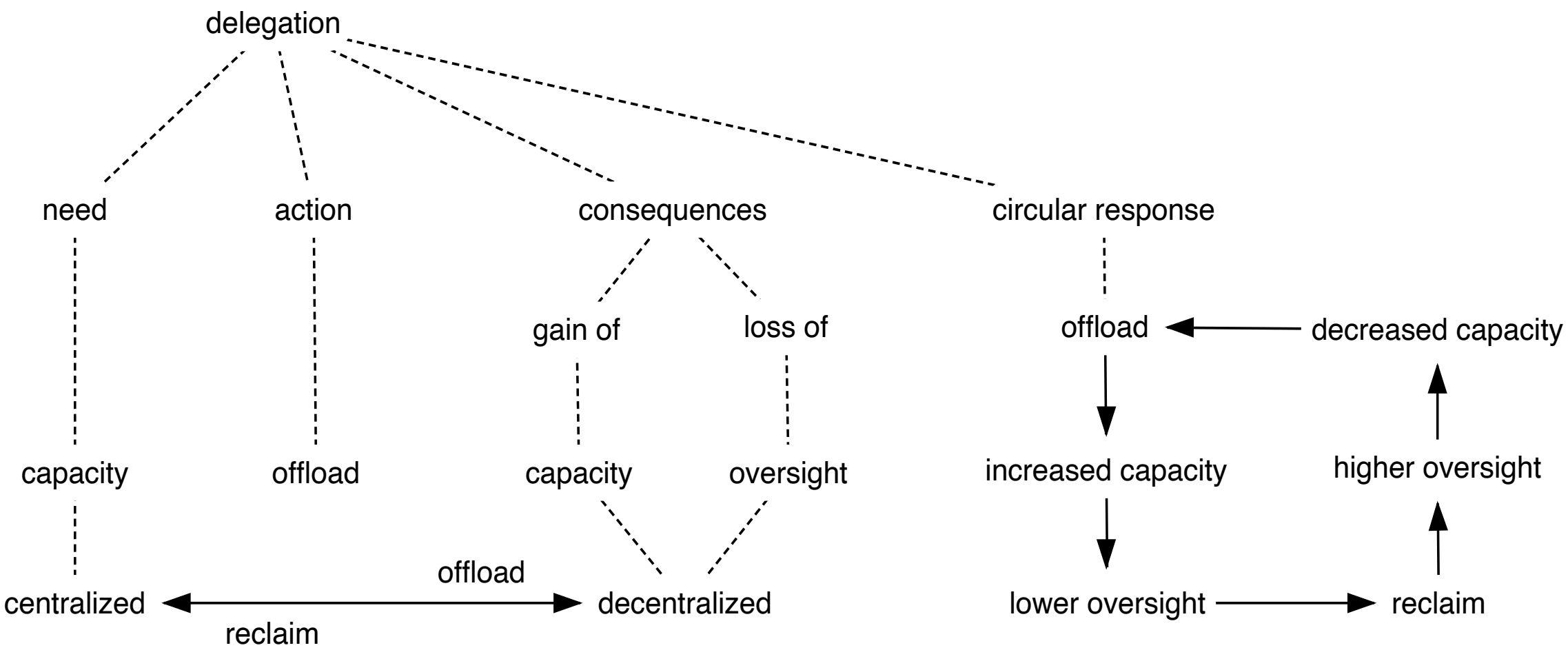

**Figure 3.** Mapping of the paradox for the delegation scenario.

Figure 4 shows the mapping of the paradox for the assistance scenario. In this scenario, the central tension is one between whole and part. The part is what the AI assists with.

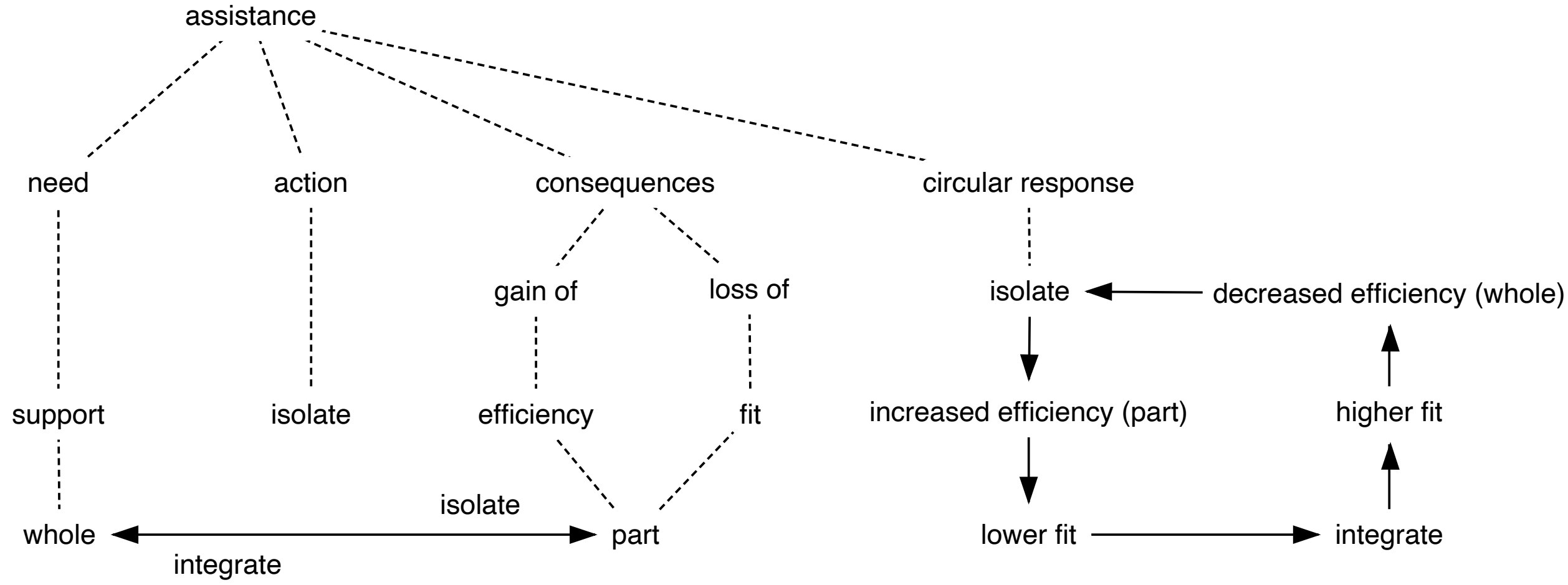

**Figure 4.** Mapping of the paradox for the assistance scenario.

Figure 5 shows the mapping of the paradox for the co-creation scenario. At the heart of this scenario is the tension between self and other. The role of the other is played by the AI.

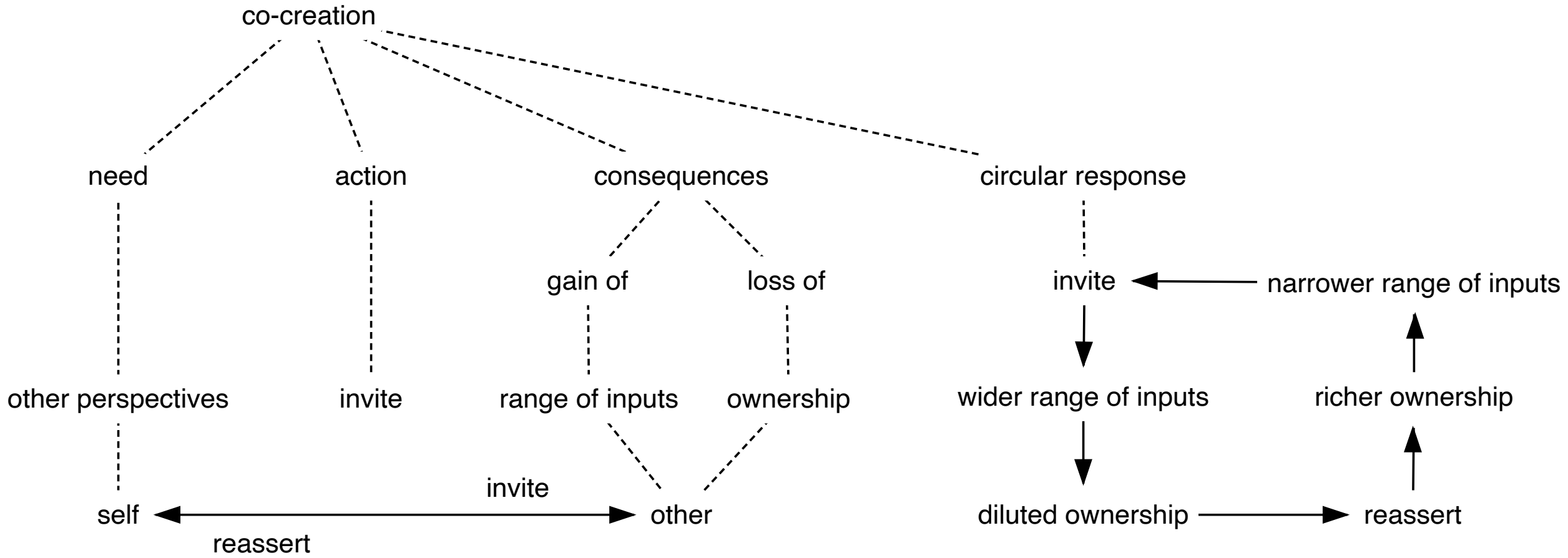

**Figure 5.** Mapping of the paradox for the co-creation scenario.

## 4.3 From paradoxes to patterns

Mapping a paradox to a pattern involves the following steps:

1. Identify the scenario (from the design space)
2. State the context (scenario-specific condition, containing paradox)
3. Name the paradoxical tension (interdependent forces that are in perpetual tension)

4. Write the problem statement (use both/and framing, not resolvable)
5. Explain the forces (effect of each force's requisite pull)
6. Describe the solution (name the action/reaction pair, design space configuration)
7. State the resulting context (trace the circular response)
8. Identify related paradoxes (nested paradoxes)

Table 2 shows the mapping for the instruction paradox.

**Table 2.** Example of applying the eight steps for the instruction paradox

| | Step | Description |
|---|---|---|
| 1 | Scenario | Instruction |
| 2 | Context | The knowledge needed to perform the task is still tacit |
| 3 | Paradoxical tension | Tacit vs. explicit |
| 4 | Problem statement | Codification requires structure, but structure reduces sensitivity<br>It sharpens and dulls your focus |
| 5 | Forces | Tacit knowledge is difficult to externalize<br>Probing involves translating partial understanding<br>Recognizing deviations requires sensitivity<br>Structure can mute aspects of a task |
| 6 | Solution | Iteratively probe the AI and reflect on its responses |
| 7 | Resulting context | Probing increases sensitivity, loosens structure<br>Reflection tightens structure, decreases sensitivity |
| 8 | Nested paradoxes | Managing tacit/explicit enables delegation (centralized/decentralized), assistance (whole/part), and co-creation (self/other) |

# 5 Patterns

The essence of each scenario in Figure 1 can be captured as a pattern. A pattern is a recurring solution to a common problem in a context (Alexander et al., 1977). The patterns will be documented in the Alexandrian format. In this format, the context, problem and solution, and consequences are visually separated by diamonds. The problem and solution statements are also highlighted.

The patterns do not deal with the decision when to use AI. The assumption is that the decision to engage with AI has already been made, and the challenge is how to collaborate with AI. The patterns also assume that the user is a domain expert, somebody who can, therefore, assess the quality of the output produced by an AI. The patterns, thus, do not directly apply to novices or learners.

## 5.1 Instruction

You are a domain expert. You have decided to use AI to support a task in your field. However, the knowledge required to perform the task is still tacit, that is, locked in your head.

**Codifying tacit knowledge into instructions for an AI requires you to structure your thinking, but structure comes at the cost of what it fails to anticipate. It both sharpens and dulls your focus.**

Tacit knowledge is knowledge that humans acquire through experience and that is difficult to fully externalize (Polanyi, 1962). Polanyi (1962) writes that “there are things that we know but cannot tell.” Prompting (either a single prompt or a conversation) involves translating partial or inarticulate ideas about how to perform a task into a form that can be manipulated by an AI (Dalsgaard, 2025). By articulating an idea and assessing the AI’s outputs, you can then revise your own understanding of the task.

Deviations in the AI's responses can offer insights into its behavior and help you create a mental model of the AI. Weick & Sutcliffe (2015), in their study of high-reliability organizations, refer to this as sensitivity to operations, the need to monitor a system’s “messy” reality in order to respond promptly to unexpected deviations. These steps, articulating your understanding of a task and monitoring the AI’s responses can be understood as probing a system and reflecting on what you learn from it.

AI systems act as “malleable objects-to-think-with” (Granata, 2025), which you can shape through iterative prompting. As you probe and reflect, you externalize your thinking and infer how the AI operates. Your perception is also shaped by what you pay attention to. Prompting can highlight certain aspects of a task while muting others (Dalsgaard, 2025). Initially, the structure of a prompt should be loosely defined so that it can be adjusted as the user learns more about the AI’s actual behavior.

Therefore,

**Iteratively probe the AI and reflect on its responses, so that any deviations between the AI’s responses and your own judgment can be identified and corrected.**

This is the low-autonomy, low-initiative scenario. With each prompt, you ask the AI to perform specific tasks (low autonomy) and the interaction remains under your direct control (low initiative).

The workflow in Figure 6 shows a human (blue) interacting with a chatbot (orange) by asking questions. The chatbot, in turn, may search a database (green) to provide answers. The human probes the chatbot by structuring questions and then reflects on the answers.

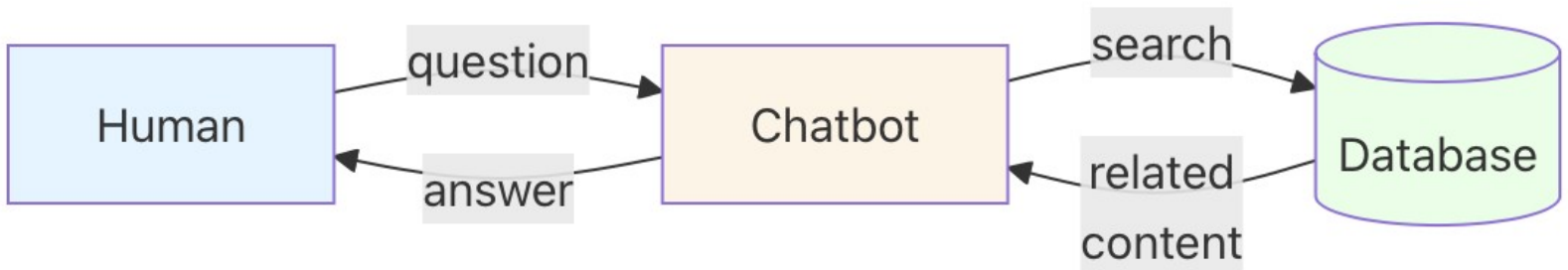

**Figure 6.** Example workflow for Instruction (blue = human, orange = agent, green = tool).

The example in Table 3 illustrates how probing and reflecting leads to an improved understanding of the task you want to achieve and how to structure the task for an AI. The first prompt is ambiguous; the word “pattern” can have multiple meanings. In the second prompt, the user defines the specific way in which the term “pattern” is used by referring to Alexander et al.’s (1977) definition. The third prompt instructs the AI to ask the user questions about the pattern they want to write.

Judgment is the part of tacit knowledge that is most difficult, if not impossible to externalize.

**Table 3.** Example of using Instruction (the example conversation is fictitious, but it was tested with a vanilla version of Perplexity that did not keep any memory of my previous interactions)

> Help me write a pattern.

[This is too vague. Since you didn’t specify what you meant by a “pattern,” the system might come back with a knitting pattern or a scale pattern for a guitar, rather than a design pattern.]

> Pattern as in a common solution to a recurring problem in a context.

[This is more specific. However, you are still likely only to get examples of patterns. Depending on the AI you are using, it may or may not also ask you about the pattern you want to create. If not, you need to instruct it explicitly to ask you questions to collect information about the pattern.]

> Ask me questions to collect information about the pattern.

[The AI now asks questions until it is has enough information to write the pattern.]

With each round of feedback and corrections (reflect) to your prompts (probe), the AI’s responses become more aligned with your expectations. However, your instructions can never fully capture the whole of your tacit knowledge. That is the realm of human judgment.

Codifying tacit knowledge requires structure. With each cycle, the instructions make more of your intent and tacit task-related knowledge explicit by making the prompts more structured. However, this also makes you less sensitive to noticing any misalignments or surprising outcomes.

However, when you impose structure too early, you give up on the open, exploratory quality of prompting. Over-structuring could limit the speculative and interpretive negotiation (Dalsgaard, 2025) through which you learn, replacing it with efficient but shallower generation of output.

Depending on the type of task you want to be sensitive of how structuring your prompts impacts your ability to do complete it. If you are using AI for functional, repetitive tasks, these are going to be more clearly defined already, and you can quickly define their structure. For exploratory, reflective tasks, however, imposing structure too early can result in the loss of the very openness and ambiguity that make prompting generative (Dalsgaard, 2025). As Dalsgaard (2025) writes, the goal for those tasks is not to create a perfect prompt on the first try, but to learn what the right question should be.

Once a large subtask has been sufficiently codified through probing and reflection and is ready to be delegated as a whole to AI, move on to Delegation. If the codified task is a smaller, incidental task, then consider Assistant. Finally, if remaining exploratory is important then proceed on to Co-Creation for tasks in which the overall structure of human-AI interaction can be codified and assigned to a team of agents while leaving space for iterative exploration. While Instruction is like a lone inventor, Co-Creation is more like an R&D department. Transitioning to Delegation, Assistance, or Co-Creation leads to nesting a new paradoxical tension within the tacit vs explicit paradox.

## 5.2 Delegation

Your work involves self-contained tasks that can be delegated. You used Instruction to develop a good understanding of how to structure these tasks so they can be performed by AI.

**Offloading work to AI allows you to focus on higher-level tasks and increases the amount of work you can complete, but it comes at the cost of disengaging with the output.**

Therefore,

**Selectively verify the output of delegated tasks. When the output does not align with your expectations, reclaim authority over the output and review the task structure.**

This is the high-autonomy, low-initiative scenario. The AI operates independently but within set boundaries (high autonomy), and waits for you to initiate the task (low initiative).

Figure 7 shows a chain of tools and agents[†] that a researcher might use to extract metadata from papers and analyze the abstracts (e.g., to identify the research problem). The role of the human is to specify the workflow and verify the results, if needed. No human is involved in executing this workflow.

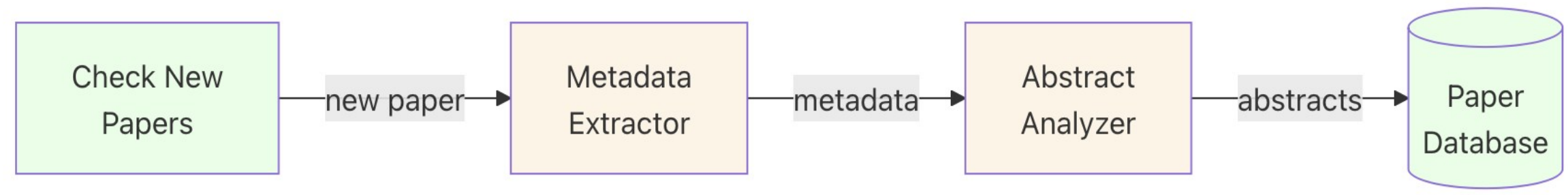

**Figure 7.** Example workflow for Delegation (blue = human, orange = agent, green = tool).

## 5.3 Assistance

Your work includes both higher-level cognitive and lower-level routine tasks. You used Instruction to help you structure those lower-level tasks so that they can be performed by AI.

**Isolating a part of your work and offloading it to AI increases the efficiency of that part, but it comes at the cost of losing fit with the whole task.**

Therefore,

† Agents are entities that couple perception, reasoning, and acting; they act in an environment that often includes other agents (Poole & Mackworth, 2023). An agent system consists one or multiple agents (Weiss 2001).

**Review the output of the part offloaded to AI against the goal of the whole task. Integrate it with the surrounding context to restore fit between part and whole.**

This is the low-autonomy, high-initiative scenario. Although the AI is only given a supporting task to perform (low autonomy), it initiates execution based on monitoring the context (high initiative).

The workflow in Figure 8 shows a chain of agents embedded in an editor that provide feedback on the claims identified in the most recent set of changes made by the human user.

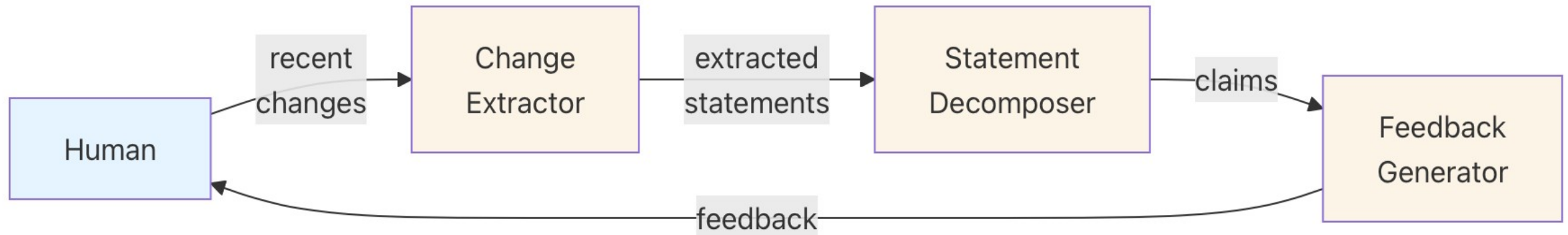

**Figure 8.** Example workflow for Assistance (blue = human, orange = agent, green = tool).

## 5.4 Co-Creation

Your work requires multiple, diverse perspectives. You used Instruction to help you understand how you can collaborate with AI as a peer helping you shape the output.

**Inviting the AI's perspective into your work widens the range of inputs you can draw on, but it comes at the cost of diluting your ownership of the work.**

Therefore,

**Invite the AI's distinct perspective, then reassert your own judgment, voice, and intent to maintain ownership of the outputs as the work evolves.**

This is the high-autonomy, high-initiative scenario. Here, the AI acts as a partner. It operates autonomously with broad authority (high autonomy), and initiates actions (high initiative).

Figure 9 shows a workflow for conducting a thematic analysis with touchpoints for the human user to provide input and feedback. It contains agents for extracting codes, identifying themes from the codes, and defining the final themes. After the agents identified themes, the user can revise the themes until they are satisfied with the result. This is where the human asserts ownership over the output.

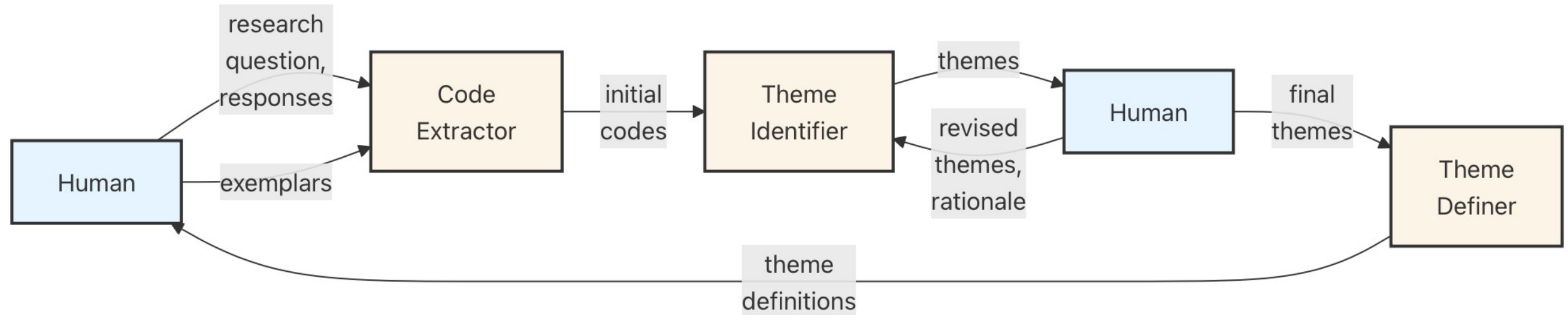

**Figure 9.** Example workflow for Co-Creation (blue = human, orange = agent, green = tool).

# 6 Conclusion

This paper makes two contributions. It describes patterns for human-AI interaction and documents a process for going from tensions to paradoxes and from paradoxes to patterns. The process expands on the process introduced in Weiss (2026). Its goal is twofold: first, to give a more dynamic view of the paradox, and second, to derive the pattern description from the model of the paradox.

Future work includes testing the process on a larger collection of patterns. Another line of research will be to model paradoxes formally using distinctions, building on prior work on modeling patterns as distinctions (Weiss, 2025), but applying distinctions to model paradoxical tensions as re-entrant forms, instead, and to express the nesting of paradoxes that results from applying patterns in sequence.

# Acknowledgments

I thank Mohammad Daud Haiderzai for his helpful comments during the shepherding phase. I would also like to express my gratitude to the participants of my Writers' Workshop at PLoP.